\documentclass{article}

\usepackage{arxiv}

\usepackage[utf8]{inputenc} 
\usepackage[T1]{fontenc}    
\usepackage{hyperref}       
\usepackage{url}            
\usepackage{booktabs}       
\usepackage{amsfonts}       
\usepackage{nicefrac}       
\usepackage{microtype}      
\usepackage{lipsum}		
\usepackage{graphicx}
\usepackage{multirow}
\usepackage{adjustbox}
\usepackage{xcolor}
\usepackage{graphicx}

\title{A multimodal approach for\\ multi-label movie genre classification
}

\author{Rafael B. Mangolin \\
        Department of Informatics \\
        State University of Maring\'a \\
        Maring\'a -- Paran\'a -- Brazil \\
	    \texttt{rbmangolin@gmail.com} \\ \and
        Rodolfo M. Pereira\\
        Federal Institute of Paran\'a and \\
        Pontifical Catholic\\ 
        University of Paran\'a\\
        Pinhais - Paran\'a - Brazil\\
        \texttt{rodolfomp123@gmail.com} \\  \and
        Alceu S. Britto Jr.\\
        Pontifical Catholic\\ 
        University of Paran\'a\\
        Curitiba - Paran\'a - Brazil\\
        \texttt{alceu@ppgia.pucpr.br} \and
        Carlos N. Silla Jr.\\
        Pontifical Catholic\\ 
        University of Paran\'a\\
        Curitiba - Paran\'a - Brazil\\
        \texttt{carlos.sillajr@gmail.com}  \and
        Valéria D. Feltrim \\
        Department of Informatics \\
        State University of Maring\'a \\
        Maring\'a -- Paran\'a -- Brazil \\
	    \texttt{valeria.feltrim@gmail.com} \and
        Diego Bertolini\\
        Federal Technological\\ 
        University of Paran\'a\\
        Campo Mour\~ao - Paran\'a - Brazil\\
        \texttt{diegobertolini@utfpr.edu.br}\\ \and
        Yandre M. G. Costa \\
        Department of Informatics \\
        State University of Maring\'a \\
        Maring\'a -- Paran\'a -- Brazil \\
	    \texttt{yandre@din.uem.br} \\
}

\begin{document}
\maketitle

\begin{abstract}
Movie genre classification is a challenging task that has increasingly attracted the attention of researchers. The number of movie consumers interested in taking advantage of automatic movie genre classification is growing rapidly thanks to the popularization of media streaming service providers. In this paper, we addressed the multi-label classification of the movie genres in a multimodal way. For this purpose, we created a dataset composed of trailer video clips, subtitles, synopses, and movie posters taken from 152,622 movie titles from The Movie Database (TMDb)\footnote{https://www.themoviedb.org}. The dataset was carefully curated and organized, and it was also made available\footnote{https://drive.google.com/file/d/1X9siuhXAsCdQljDjjs9PbqWXQ5Z72FOV/view?usp=sharing} as a contribution of this work. Each movie of the dataset was labeled according to a set of eighteen genre labels. We extracted features from these data using different kinds of descriptors, namely Mel Frequency Cepstral Coefficients (MFCCs), Statistical Spectrum Descriptor (SSD), Local Binary Pattern (LBP) with spectrograms, Long-Short Term Memory (LSTM), and Convolutional Neural Networks (CNN). The descriptors were evaluated using different classifiers, such as BinaryRelevance and ML-kNN. We have also investigated the performance of the combination of different classifiers/features using a late fusion strategy, which obtained encouraging results. Based on the F-Score metric, our best result, 0.628, was obtained by the fusion of a classifier created using LSTM on the synopses, and a classifier created using CNN on movie trailer frames. When considering the AUC-PR metric, the best result, 0.673, was also achieved by combining those representations, but in addition, a classifier based on LSTM created from the subtitles was used. These results corroborate the existence of complementarity among classifiers based on different sources of information in this field of application. As far as we know, this is the most comprehensive study developed in terms of the diversity of multimedia sources of information to perform movie genre classification.

\end{abstract}

\section{Introduction}
\label{intro}

Streaming media services have grown steadily over the past decade, mainly due to the consolidation of video on demand as a practical and comfortable way of allowing consumers to have access to films, series, documentaries, and so on. Some giant companies (e.g.: Netflix\texttrademark, Hulu\texttrademark, Amazon\texttrademark prime video, YouTube\texttrademark and Facebook\texttrademark watching)\footnote{https://www.tapjoy.com/resources/video-streaming-industry-growth/} are rapidly gaining ground in this market, as they offer exclusive content through agreements with the movie industry and integrate other products and services that serve consumers' interests more comprehensively.

In parallel, the multimedia retrieval research community has been devoting efforts to assess new methods and techniques that seek to properly explore and retrieve movies based on data sources normally available with movie titles. In this sense, many studies focus on trailer content (audio and/or image), synopsis texts, posters images, and so on \cite{zhou2010movie, portolese2018use, simoes2016movie, huang2012movie}. Although some of these studies have adopted multimodal approaches, none of them combines data sources as proposed in this work.

In this study, we addressed the movie genre classification using multimodal classifiers based on audio and images from trailers (i.e. audio and video frames), subtitles, posters, and synopses in a multi-label scenario. 
First, we obtained representations from each modality using both handcrafted and non-handcrafted (i.e. obtained using representation learning) features, totaling 22 types of features. Then, we create multi-label classifiers for each feature using the multi-label classification algorithms Binary Relevance and ML-kNN. The predictions provided by the classifiers were combined using different late-fusion strategies (sum rule, product rule, mean rule, and max rule). Aiming to prevent issues related to the dataset imbalances, we also perform experiments with the resampling techniques ML-SMOTE, MLTL and a combination of both. Besides, we have tried compressive sampling, a promising strategy employed aiming to provide dimensionality reduction. It was particularly tested with some very large representations created from textual data sources.

To our knowledge, this is the first study to combine all of these data sources for movie genre classification. \cite{cascante2019moviescope} have presented a dataset which include all the sources of information used here, but they did not experiment with all of them. 

To evaluate our proposal, we carefully curated and organized a dataset based on movie titles taken from The Movie Database (TMDb). The dataset contains 10,594 movies labeled in 18 genres. For each movie the following data was collected: the video trailer, the poster, the synopsis, and the subtitle. The movies' trailers were collected from YouTube\texttrademark, and the subtitles were retrieved from OpenSubtitles. We have chosen subtitles and synopsis in English because they are by far the most widely used in the world. This dataset was used in our experimental protocol and made available to the community, thus allowing other researchers to evaluate other protocols and properly compare the obtained results with those presented here. 

The main contributions of this paper are: 
\begin{itemize}
    \item the development of a multimodal database for movie genre classification already made available to the community, which allows to replicate the results, and to compare to other approaches;
    \item a multimodal approach to perform the multi-label genre classification;
    \item evaluation of different representations, including hand-crafted and deep learning techniques;
    \item evaluation of the performance of late fusion techniques to combine different data sources models.
\end{itemize}



The remaining of this paper is organized as follows: in Section \ref{sec:works} we summarize some of the related works regarding movie genre classification and multimodal multimedia classification. The dataset carefully curated to support the development of the experiments described here is presented in Section \ref{sec:dataset}. In Section \ref{sec:method} we present the multimodal approach proposed in this study. In Section \ref{sec:results} we present our experimental results. The discussion of the results is made in Section \ref{sec:discussion}. And finally, Section \ref{sec:conclusion} brings concluding remarks and directions for future works.

\section{Related works}
\label{sec:works}

Several researchers have devoted efforts towards the development of new approaches to movie genre classification. In their studies, they have explored different kinds of data as source of information, such as movie posters, subtitles, trailers, and synopses. The 
23
 features captured from these sources are usually used to feed classifiers induced by machine learning algorithms. In this section, we present some of these works in chronological order.

\cite{brezeale2006using} used bag-of-words (BOW) taken from closed captions and bag-of-visual-features (BOVF) taken from frames of video clips to classify movie genres. BOW features were extracted after the removal of stop words \cite{frakes1992information} and stemming \cite{porter1980algorithm} from the closed captions. A histogram was computed for each movie using Discrete Cosine Transform (DCT), which was applied to the first frames of each scene of the movie. The histograms were submitted to the k-means algorithm to generate the BOVF. Both BOW and BOVF were evaluated in experiments with SVM classifiers on a multi-label dataset of 81 movies from the MovieLens project labeled with 18 genres. The authors performed a binary classification creating an SVM for each genre. The result was obtained using the mean accuracy of the classifiers. The mean accuracy obtained using BOW and BOVF was 89.71\% and 88.48\%, respectively. 
 
\cite{zhou2010movie} used GIST visual descriptor \cite{oliva2001modeling}, CENTRIST \cite{wu2008place}, and W-CENTRIST to extract features for movie genre classification. The used descriptors emphasize the textural content of the frames (GIST), the intensity pattern between the pixels (CENTRIST), and the chrominance patterns between the pixels (W-CENTRIST). The used dataset contained 1,239 movie trailers classified into four main genres, namely action, comedy, drama, and horror. BOVF features were generated using k-means with a hundred centroids. The best accuracy was 74.7\%, obtained using CENTRIST features.

\cite{huang2012movie} proposed the multimodal classification of movie genre using features extracted from the audio and video content (i.e. frames) of trailers. 
The visual features were classified into two categories, temporal (i.e. time related) and spatial (i.e. frame content related), totaling 75 features. The audio features were categorized in audio intensity related, rhythm related, and timbre related. A total of 202 audio features were used, including Mel-Frequency Cepstral Coefficients (MFCC), one of the most widely used features in tasks focused on audio classification. A feature selection step was performed using the Self-Adaptive Harmony Search (SAHS) algorithm \cite{wang2010self} on each of the one-against-one SVMs employed in the classification. The used dataset was composed of 223 video trailers classified in seven genres (i.e. action, animation, comedy, documentary, drama, musical, and thriller) and the best obtained accuracy was 91.9\%.

\cite{hong2015multimodal} also proposed a multimodal approach using movie trailer content (i.e. video, audio, and tags) to perform movie genre classification. The authors used two concepts taken from the Probabilistic Latent Semantic Analysis (PLSA) \cite{hofmann2001unsupervised} to combine features extracted from text (i.e. social tags), audio, and image. The dataset used in the experimental protocol was composed of 140 movie trailers distributed in four classes (i.e. action, biography, comedy, and horror), and social tags obtained via social websites. The proposed approach achieved an accuracy of 70\% using the early fusion of audio, video, and text.
 
\cite{fu2015fast} combined features extracted from movie posters and synopses to perform movie classification. From the posters, features related to color, texture, shape, and the number of faces depicted in the image were extracted. From the synopses, BOW features were extracted after preprocessing the texts for stop words removal and stemming \cite{porter1980algorithm}. The dataset used contained synopses and posters from 2,400 movies taken from TMDb\footnote{https://www.themoviedb.org/} classified into four genres (i.e. action, comedy, romance, and horror). Two SVM classifiers were created, one using features obtained from posters, and other with features taken from synopses. The best accuracy rate achieved was 88.5\%, when the scores obtained with both classifiers were fused.

\cite{simoes2016movie} used deep learning to classify movie genres. Features were extracted from each trailer frame using a CNN. Then, for each trailer scene, a feature vector with the average values of its frames was computed and submitted to the k-means algorithm to generate a BOVF. Audio features were also extracted from the trailers using MFCC. Experiments were performed using the LMTD-4 dataset, which is a subset taken from the Labeled Movie Trailer Data (LMTD), created by the authors. LMTD-4 is composed of 1,067 trailers classified into four genres (i.e. action, comedy, drama, and horror). The best accuracy rate was 73.75\%, obtained using SVM with BOVF features, audio features, and the weighted prediction generated by the CNN.

\cite{wehrmann2017movie} used movie trailers to perform multi-label genre classification, exploring features taken both from audio and video contents. The authors used CNN with the Convolution Through-Time (CTT) module, which organizes the features taken from each frame considering their respective positions on the sequence as a whole, thus forming a matrix. Aiming at identifying the spatial features of the trailer, this matrix is processed by a convolutional layer. The authors proposed five different models. In three of them (CTT-MMC-A, CTT-MMC-B, and CTT-MMC-C), based on video content, the configuration of the classification layers was changed; one of them (CTT-MMC-S) uses audio content; and in the fifth model (CTT-MMC-TN), the results obtained by the networks trained using audio are fused with the results obtained by the networks using video (CTT-MMC-TN). Experiments were conducted on a subset of the LMTD dataset, LMTD-9, which is composed of trailers from 4,007 movies, labeled into nine genres.
The best result was obtained using the model CTT-MMC-TN, achieving an AUC-PR (Area Under Precision--Recall Curve) of 74.2\%.

\cite{portolese2018use} performed multi-label movie genre classification based on features extracted from the synopses. The authors evaluated 19 feature sets based on Term Frequency-Inverse Document Frequency (TF-IDF) features, as well as different models of word and document embeddings. The dataset used in the experiments was composed of 12,094 movie synopses written in Brazilian Portuguese collected from TMDb \footnote{https://www.themoviedb.org/} and classified into 12 genres. Four different classifiers were used (i.e. Multilayer Perceptron (MLP), Decision Tree, Random Forest, and Extra Trees), and the best result (i.e. 54.8\% of F-Score) was obtained using an MLP classifier fed with TF-IDF features based on 3-grams with dimensionality equal to 1,000. In \cite{portolese2019exploring}, the authors extended their dataset to 13,394 movies (still with Portuguese synopses only) classified in 18 genres, and the groups of textual features. They also experimented with an oversampled version of the dataset. The best result for the original dataset was achieved by a TF-IDF based classifier, presenting an average F1-score of $0.478$, while the best result for the oversampled dataset, an average F1-score of $0.611$, was achieved by a combination of several feature groups. 

\cite{cascante2019moviescope} carried out a large scale investigation comparing the effectiveness of visual, audio, text, and metadata-based features for the prediction of high-level information about movies, including genre classification. The dataset used in this work was composed of 5,027 movies classified with 13 labels. They presented a multimodal dataset with trailer, text plot, poster, and other kinds of metadata. They concluded that the use of features obtained from text (i.e. fastText) and video (i.e. fastVideo) are better suited for a holistic classification task than the well-known LSTM-based representations. They achieved a mean average precision of 68.6\% performing late fusion. 

Despite focusing on a different kind of application, music genre classification, the work of \cite{oramas2017multi} was also important for this study since the authors used deep learning techniques with multimodal data. For classifying music genre, representations obtained using deep architectures were learned from audio tracks, text reviews, and cover art images. These representations were evaluated both individually and together. The obtained results showed that the combination of representations from different modalities performs better than any of the modalities in isolation, indicating a complementarity using multimodal classification. 

Table \ref{tab:works} presents a summary of the studies described in this section, focusing on the most important characteristics of the works related to the task investigated here (i.e.: Author/year, source of information, content descriptor, ML technique, and recognition rate). It is worth mentioning that the purpose of this table is to summarize and not to compare, as it is not possible to make a fair comparison among the works described in this section, since they were developed using different databases and sources of information.

\begin{table}[htbp!]
\centering
\caption{Summary of the works described in this section.}
\label{tab:works}
\begin{tabular}{p{2cm} p{2cm} p{2,5cm} p{2cm} c}
\hline
\multicolumn{1}{c}{\begin{tabular}[c]{@{}c@{}}\textbf{Author/}\\\textbf{year}\end{tabular}} & \multicolumn{1}{c}{\begin{tabular}[c]{@{}c@{}}\textbf{Source of}\\ \textbf{information}\end{tabular}} & \multicolumn{1}{c}{\begin{tabular}[c]{@{}c@{}}\textbf{Content}\\ \textbf{descriptor}\end{tabular}} & \multicolumn{1}{c}{\begin{tabular}[c]{@{}c@{}}\textbf{ML}\\ \textbf{technique}\end{tabular}} & \multicolumn{1}{c}{\begin{tabular}[c]{@{}c@{}}\textbf{Recognition}\\ \textbf{rate}\end{tabular}}
\\ \hline
\cite{brezeale2006using}* & Subtitles and movies & BOW and BOVF & SVM & 89.71\%$^{1}$ \\

\cite{zhou2010movie}* & Frames &  GIST, CENTRIST and W-CENTRIST & \textit{k-means} & 74.7\%$^{1}$ \\

\cite{huang2012movie} & Trailers & Structural video descriptors, MPEG-7, RMS, MFCC and LPC & SAHS and SVM & 91.9\%$^{1}$ \\

\cite{hong2015multimodal} & Tags and trailers & BOW, BOAF and BOVF & PLSA & 70\%$^{1}$ \\ 

\cite{fu2015fast} & Posters and synopsis & BOW and visual descriptors & SVM & 88.5\%$^{1}$ \\

\cite{simoes2016movie} & Trailers & BOVF and MFCC & SVM & 73.75\%$^{1}$ \\

\cite{wehrmann2017movie}* & Trailers & CNN & CNN & 72.4\% $^{2}$ \\

\cite{portolese2018use}* & Synopsis & TF-IDF and  \textit{word embeddings} & Decision trees and MLP & 54.8\%$^{3}$ \\

\cite{cascante2019moviescope}*  & Trailers, movie plots, posters and metadata & CNN and LSTM & CNN and LSTM & 68.6\%$^{4}$\\

\cite{portolese2019exploring}* & Synopsis & Several textual features & Binary Relavance with Decision trees & 61.1\%$^{3}$ \\
\hline
\multicolumn{5}{l}{$^{1}$ Accuracy}\\
\multicolumn{5}{l}{$^{2}$ AUC-PR}\\ 
\multicolumn{5}{l}{$^{3}$ F-Score}\\
\multicolumn{5}{l}{$^{4}$ mAP}\\
\multicolumn{5}{l}{* multi-label classification}\\
\end{tabular}
\end{table}

\section{Dataset}
\label{sec:dataset}

We carefully curated and organized a multimodal dataset of movie titles to the development of the experimental protocol described in this work. The data was collected from \textit{The Movie Database} (TMDb), a database created in 2008 that contains freely available data regarding movies and TV series from all over the world. Among the most remarkable aspects related to the TMDb, we highlight: i) different data sources (i.e. synopsis, poster, trailer, and subtitle) are available for each movie title; and ii) the movies are labeled with one or more genres, i.e., in a multi-label way.

The creation of this dataset was inspired by the one described by \cite{portolese2019exploring}, which the authors also used a subset of TMDb, but composed only of Portuguese synopses from 13,394 movies. We used that subset of titles as a starting point, however, we focused on retrieving the text data sources (i.e. synopsis and subtitle) in English rather than in Portuguese. We invested plenty of effort to expand it, aiming at composing a new dataset of movies that includes synopsis in English, subtitles, video trailers and poster for all titles. 

Our final dataset is composed of data from 10,594 movies titles and each title have:  the poster (in image format), the subtitle (in text format), the synopsis (in text format), and the movie trailer clips (which allow the use of both audio and video content, as discussed in subsections \ref{subsec:audio} and \ref{subsec:image}, respectively). It is worth mentioning that some of the data were not available for all titles of the initial subset, and that is why our final dataset is smaller than the initial one. All data collection was finished in July 2019.

To collect the subtitles, we used a two steps protocol. In the first step, the name, debut year, and ID of the movie in the IMDB\footnote{https://www.imdb.com/} website were obtained using the TMDb API\footnote{https://www.themoviedb.org}. In the second step, the subtitle was retrieved from the OpenSubtitles\footnote{https://www.opensubtitles.org} website, either using the IMDB ID, when available, or searching for the movie’s title and debut year. Using this protocol, we collected the English subtitles for 11,066 movies.

We also used the TMDb API to obtain the movies’ posters and synopses. We collected posters for 13,047 movies and the English synopsis for 12,320 movies.

We collected the movies’ trailers from YouTube\footnote{https://www.youtube.com/}. To identify which video was related to which movie, we used two strategies: if the URL of the movie’s trailer on YouTube\texttrademark was available on TMDb, we used it to download the video; if not, we used the movie’s title and debut year to search its trailer using the YouTube API. For each movie searched, we chose the first returned result with a duration length between 30 and 300 seconds. The trailers were downloaded using a Python library. Using this protocol, we collected trailers for 13,390 movies.

After collecting all the data, we selected only the movie titles with all four data sources (i.e., English subtitles, English synopsis, Poster, and trailer), totaling 10,594 movie titles. 

\begin{figure}[!h]
    \centering
    \includegraphics[width=1.0\textwidth]{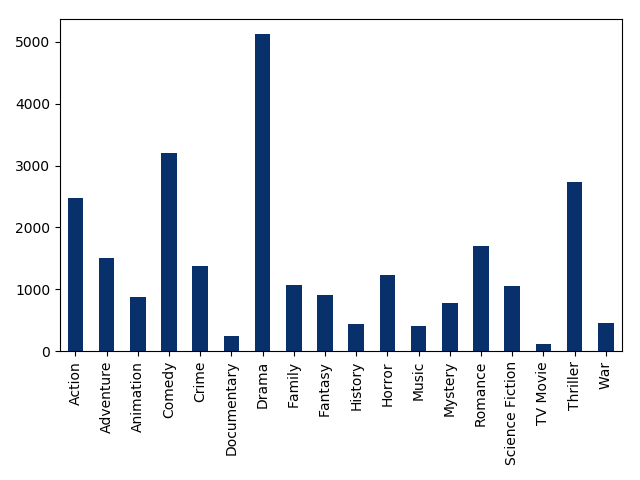}
    \caption{Number of samples per genre.}
    \label{fig:genres-count}
\end{figure}

As in \cite{portolese2019exploring}, the movies in our dataset are labeled with 18 different genre labels, namely: Action, Adventure, Animation, Comedy, Crime, Documentary, Drama, Family, Fantasy, History, Horror, Music, Mystery, Romance, Science Fiction, TV Movie, Thriller, and War. The number of labels assigned to a movie title ranges from one to seven. The most frequent label is Drama, assigned to 5,122 movies, and the least frequent label is TV Movie, which appears in only 119 movies. Figure \ref{fig:genres-count} shows the total amount of movie titles assigned to each genre label.

\begin{figure}[!h]
    \centering
    \includegraphics[width=1.0\textwidth]{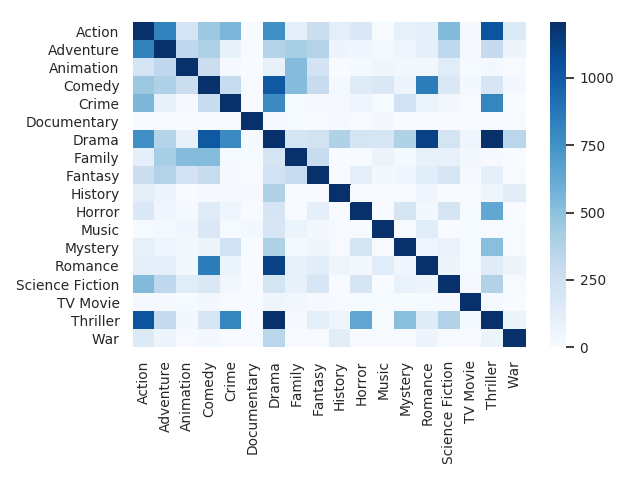}
    \caption{Labels co-occurrence matrix.}
    \label{fig:genres-heatmap}
\end{figure}

To better understand the properties of a multi-label dataset, it is interesting to observe the co-occurrence of its labels. Figure \ref{fig:genres-heatmap} shows the co-occurrence matrix for our dataset. As can be observed, Drama is the genre that more frequently co-occurs with other genres, especially Romance, Thriller, and Comedy. On the other hand, Documentary and TV Movie are the genres with less co-occurrence with other genres, which can be explained by the fact that they are assigned to only a few movies.

According to \cite{zhang2014review}, other indicators that can be observed in a multi-label dataset are the following:

\begin{itemize}
\item Label Cardinality (LCard), which corresponds to the average number of labels per example.

\item Label Density (LDen), which is the normalization of label cardinality by the number of possible labels in the label space.

\item Label Diversity (LDiv), which corresponds to the number of distinct label sets in the dataset.

\item Proportion of Label Diversity (PLDiv), which is the normalization of Label Diversity by the number of examples, indicating the proportion of distinct label sets in the dataset.

\end{itemize}

Table \ref{tab:database-metrics} presents these indicators for our dataset. As we can observe, the average number of labels per movie (LCard) is 2.426, indicating that movies in the dataset are, in average, multi-label. However, considering that there are 18 possible labels in the label space, the density of the dataset (LDen), 0.134, is low. There are 922 distinct label sets (LDiv) in the dataset. As it has 10,594 examples, the proportion of label diversity (PLDiv), 0.087, is also low. From this, we can also infer that, on average, we have about 11.49 examples per distinct label set.

\begin{table}[!htpb]
\centering
\caption{Multi-label indicators extracted from the database created for this work.}\label{tab:database-metrics}
\begin{tabular}{cc}
\hline
\textbf{Metric} & \textbf{Value} \\ \hline
LCard           & 2.426          \\
LDen            & 0.134          \\
LDiv            & 922            \\
PLDiv           & 0.087          \\ \hline
\end{tabular}
\end{table}

For the sake of transparency and also to allow other researchers to compare their experimental protocols exactly on the same dataset used here, we made our dataset available\footnote{https://drive.google.com/file/d/1X9siuhXAsCdQljDjjs9PbqWXQ5Z72FOV/view?usp=sharing}.

\section{Proposed method}
\label{sec:method}

As aforementioned in this work, we aim at exploring data from different modalities (i.e., audio, video/image, and text) to classify the genre of movie titles. Thus, we chose specific methods that could lead us to obtain the best benefit in terms of the classification performance of each of these movie title data sources.

\begin{figure}[!htpb]
    \centering
    \includegraphics[width=1.0\textwidth]{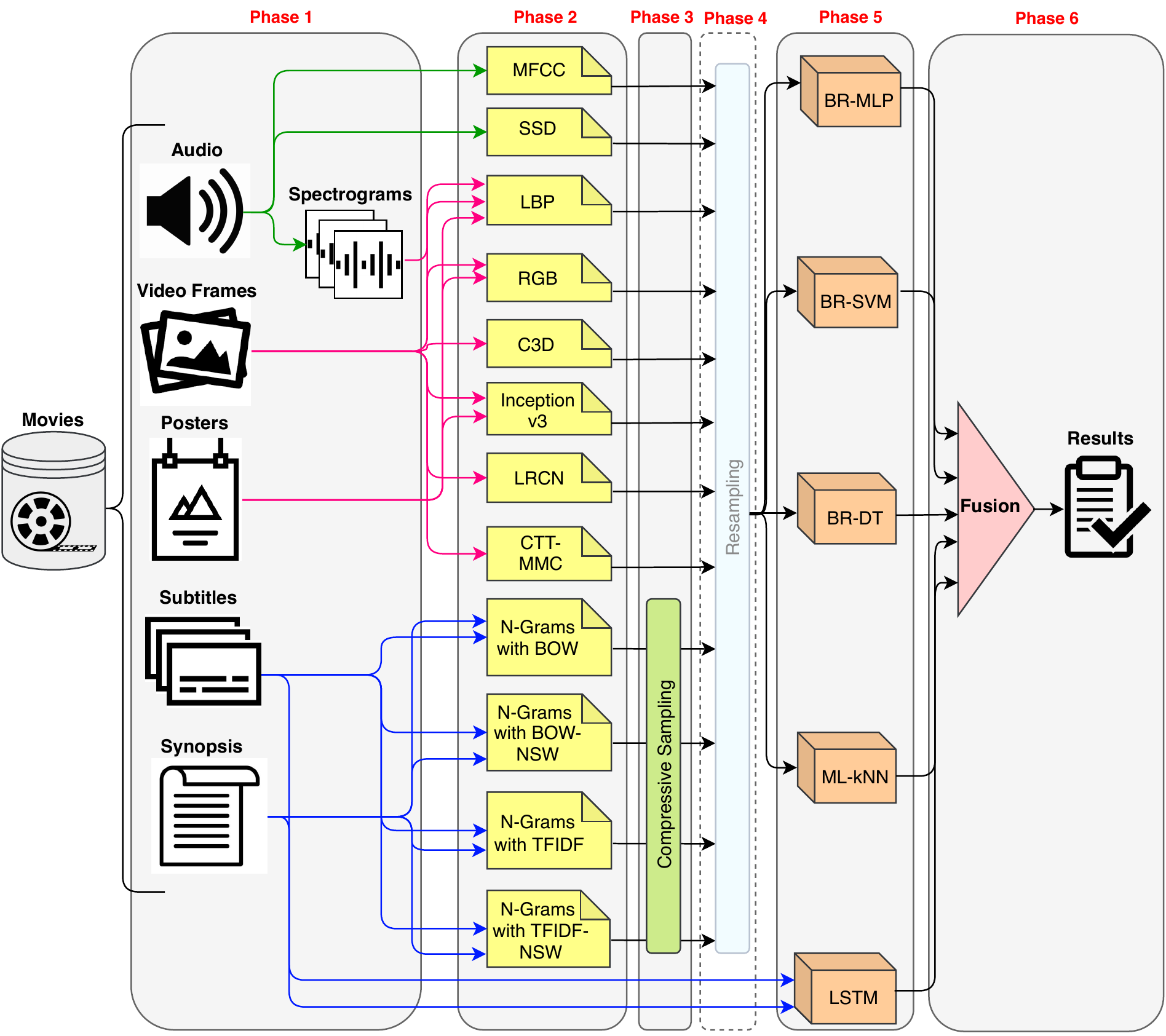}
    \caption{General overview of features and classifiers used for multimodal multi-label classification. The phase circled with a dashed rounded rectangle is optional. The different colors of the arrows represent the different types of data source used in the feature extraction phase: Green for audio sources, pink for the image sources and blue for the textual sources.}
    \label{fig:overview}
\end{figure}

To better understand the proposal of this work, Figure \ref{fig:overview} presents a general overview of it, considering: the data source preparation (Phase 1), feature extraction (Phase 2), compress (only in case of huge representations) (Phase 3), resampling (Phase 4), classification (Phase 5), and fusion of the predictions (Phase 6). It is important to note that Phase 4 is faded in Figure \ref{fig:overview} because it is an optional phase, since the original features without resampling can also be used to generate the predictions.

Before extracting the features, we performed a preprocessing stage on the data sources, represented by Phase 1 in Fig \ref{fig:overview}. In this phase, we aim to remove noises that could interfere in the classification stage. We have also applied other kinds of transformation to prepare the data for the feature extraction phase. These preprocessing was performed on different data sources such as trailers (crop and resize), audio spectrogram (crop and padding) and synopsis (removal of time marks). It is worth remembering that more details about the composition of the database have already been described in Section \ref{sec:dataset}, which can also be interpreted as part of the Phase 1.

The descriptors used to represent each type of data source (i.e. audio, video frames, poster, subtitles, and synopsis) are the following:

\begin{itemize}
    \item Audio: Mel-Frequency Cepstral Coefficients (MFCC), Statistical Spectrum Descriptors (SSD) and Local Binary Pattern (LBP);
    \item Video frames: LBP, Convolutional 3D Neural Network (C3D), Inception-v3, Long-term Recurrent Convolutional Networks (LRCN), Convolution-Through-Time for Multi-label Movie genre Classification (CTT-MMC), and RGB Feature;
    \item Poster: LBP, Inception-v3 and RGB Feature;
    \item Subtitles: Long Short-Term Memory (LSTM), and N-grams with Term Frequency–Inverse Document Frequency (TF-IDF); and
    \item Synopsis: LSTM and N-grams with TF-IDF.
\end{itemize}

As shown in Figure \ref{fig:overview}, these descriptors were used with different classifier models. Besides that, we also combined the classifiers using a late fusion strategy. 
Subsections \ref{subsec:image} and \ref{subsec:audio} describe respectively details about how the image and audio contents were explored to build the representations. 
Subsection \ref{subsec:poster} describes the poster feature extraction, and Subsection \ref{subsec:text} describes how the text content (i.e. subtitle and synopsis) was explored. 
Subsection \ref{subsec:compress} presents a summary of the representations extracted with their respective identifiers, which will be used throughout the text, and also describes the compressive sampling method, used to reduce the dimensionality of some representations obtained from the text.
Subsection \ref{subsec:resampling} describes the strategy used, as an optional step, to deal with the imbalance of the dataset.
Finally, Subsection \ref{subsec:integration} describes the algorithms used to infer the classifiers, and how the predictions of these classifiers were integrated by late fusion to get a final decision.

\subsection{Feature extraction from trailer image (Phase 2)}
\label{subsec:image}

To extract features from the selected frames of the movie trailer clips, we followed two different approaches. In the first approach, we used deep learning architectures, more specifically Convolutional Neural Networks (CNN) and Long Short-Term Memory Networks (LSTM). In the second approach, we experimented handcrafted features: we used different visual features and the $K-means$ clustering algorithm to generate the final set of features to describe the visual trailer content (i.e. Bag Of Visual Features).

Regarding the first approach, we used three techniques to extract features from the movie frames: C3D \cite{tran2015learning}, CTT-MMC \cite{wehrmann2017movie}, and LRCN \cite{donahue2015long}. We chose these architectures mainly because they perform end-to-end feature extraction and classification of spatial-temporal data. In this way, these networks process only three-dimensional data (i.e., videos), and are not suitable for processing other data sources, such as the poster or audio. The C3D architecture was designed for video processing and classification. Thus, it is equipped with convolutional and pooling operations in three dimensions to accomplish the spatial and temporal feature extraction. The CTT-MMC architecture is composed of convolutional layers disposed in two dimensions that process the video sequence frame by frame carrying out the spatial feature extraction. The features extracted from the frames are stored in matrices according to the order of the frames in the video. Then, these matrices are processed by another convolutional layer that extracts the temporal features. The LRCN architecture is similar to CTT-MMC, but instead of having a convolutional layer to extract temporal features, it uses an LSTM network\cite{hochreiter1997lstm}.

A total of 120 frames $100 \times 100$ sized were selected from each movie trailer and processed using the deep architectures. We selected the frames from each movie clip as follows. First, we discarded the first and last 5\% of the frames. We did this because these frames are often not discriminating, as their content is usually related to credits at the beginning and end of the video trailer. Then, we defined three equally distant points (i.e., start, middle, and end) and select one out of three frames from the previous 60 frames and the next 60 frames from each of those points, totaling 40 frames per point (start, middle and end) and a total of 120 frames of the trailer video as a whole.

In our second approach, we extracted features from the image content of the videos using descriptors successfully applied in other domains \cite{costa2012music} and that capture different visual attributes of the images, namely LBP (i.e. textural content)\cite{ahonen2004face} and RGB histogram (i.e. color content)\cite{gonzalez2002digital}. In this case, the number of frames used for feature extraction was defined as the lowest number of frames contained in the video trailers, which is 555. Therefore, we selected 555 frames equally distant and linearly distributed from each movie trailer. Other details related to the extraction of handcrafted features from the selected frames are the following:
\begin{itemize}
    \item Textural content: the frames were converted for grayscale images and a normalized 59-dimensional feature vector was obtained from each of them using uniform $LBP_{8,2}$ \cite{ahonen2004face}.
    \item Color content: the color content was described by a 768-dimensional feature vector made by appending the 256 sized histograms from each of the three channels (Red, Green, and Blue) of the RGB color system.
\end{itemize}

After the extraction of the handcrafted features, both datasets (i.e. textural and color feature vectors) were submitted to the $K-means$ algorithm. This algorithm assigns each frame’s feature vector to the closest centroid in the set of $n$ predefined centroids.
After some experiments varying the number of centroids from 128 to 1024, in a scale of powers of two,
we chose to use 512 centroids for LBP and 1,024 centroids for RGB histogram. After assigning the frame vectors to the centroids, the frames from each movie are grouped and a histogram is calculated for each movie group. The size of this histogram is given by the number of centroids previously defined before running the $K-means$ algorithm. In addition, this histogram is normalized by the number of selected frames (i.e. 555). The final histogram for each group, i.e., each movie, is used as the descriptor of that movie in the classification step.

\subsection{Feature extraction from audio (Phase 2)}
\label{subsec:audio}

In this section, we describe the strategy used to explore the audio content obtained by separating the sound from the image of the movie trailer video clips. 

The use of handcrafted features for the classification of audio content, captured based on a variety of descriptors, is widely present in the literature. Mel-Frequency Cepstral Coefficients (MFCC)\cite{logan2000mel}, Statistical Spectrum Descriptors (SSD)\cite{lidy2010suitability}, and Local Binary Pattern (LBP) \cite{ahonen2004face} (i.e. a powerful descriptor for the textural content of images that, in the case of sounds, can be successfully used to capture the content of spectrograms created from the audio signal) are among the most used descriptors for audio content. 
More recently, some studies have proposed the use of features learned from audio content by deep architectures \cite{oramas2017multi,salamon2017deep,pons2019randomly}. In many of these works, a deep neural network is fed by spectrogram images obtained from the original audio signal.

We explored these two approaches for extracting features from the audio content:  the first based on handcrafted features, i.e., features extracted directly from the audio signal, and the second based on features learned using deep architectures fed by spectrogram images.

For the extraction of handcrafted features, we used SSD \cite{lidy2010suitability} and MFCC \cite{logan2000mel} descriptors as they are among the most successful descriptors for audio classification tasks. Based on the human hearing system, MFCC uses the cosine discrete Fourier transform (DFT) to capture 13 features that are suitable to describe the timbre of the sound. SSD, in turn, captures information regarding audio intensity variation, also aiming at properly representing the timbre of the sound. The SSD feature vector is composed of 168 features. We also used LBP to extract textural features from the spectrograms generated from the audio. The spectrograms were created using the SOX\footnote{http://sox.sourceforge.net/} library and they are also available to other researchers as part of our data set\footnote{https://drive.google.com/file/d/1X9siuhXAsCdQljDjjs9PbqWXQ5Z72FOV/view?usp=sharing}.

In the second approach, we explored the spectrograms using the Inception-v3, a CNN architecture. Although representation learning techniques had already been proposed by \cite{Bengio2013}, the use of these techniques was reinforced thanks to the use of transfer learning using IMAGENET \cite{imagenet}. In this sense, \cite{inception-v3} proposed the Inception-v3. This architecture proved to be more robust than other architectures, presenting low error rates in the ILSVRC-2012 benchmark\footnote{http://image-net.org/challenges/LSVRC/2012/}. It also presented better results than previous architectures, such as GoogleLeNet \cite{Szegedy_2015_CVPR}, PReLU \cite{He_2015_ICCV}, and VGG \cite{SimonyanZ14a}.

To allow the training of the Inception-v3, we cropped the spectrogram images. To standardize the spectrograms width, we took the area of the spectrogram that corresponds to the 30 seconds placed in the middle of the audio clip. In the case of audio clips with less than 30 seconds duration, we have used zero padding to fill the images and keep their size in the standard. After the training of the Inception-v3, we used the 2,048 weight values of the penultimate layer of the net as feature. Before we extract the feature, we applied transfer learning using the weights of an Inception-V3 trained on the IMAGENET Dataset \cite{imagenet}.

\subsection{Feature extraction from poster (Phase 2)}
\label{subsec:poster}

As with trailer frames and audio, we explored the content of movie posters images using both handcrafted and non-handcrafted descriptors. 

In the handcrafted scenario, we decided to employ two strategies widely used to describe the textural and the chromatic content of images.
For texture description, we used LBP, which was applied for the poster image as a whole. The posters were firstly converted to grayscale images before the LBP extraction, which was accomplished using eight neighbors with radius two (i.e. $LBP_{8,2}$), resulting in a 59-dimensional feature vector \cite{ahonen2004face}. The LBP vectors were normalized before performing the classification.

Similar to the protocol applied to the frames of the trailer, we used color histograms to capture the chromatic content of the posters. In this sense, the occurrences of each intensity level on each color channel of the RGB (i.e. Red, Green, and Blue) color space were summed up to build histograms. Thus, a 768-dimensional histogram  (i.e. 3 $\times$ 256 = 768) was created and used as the feature vector to describe the chromatic content of the posters. As with LBP, these vectors were also normalized before classification.

In the non-handcrafted scenario, we used the Inception-v3 for extracting features from the poster images. The choice of this particular architecture was motivated by the reasons previously presented in subsection \ref{subsec:audio}. In this case, the network was fed with the image of the poster as a whole. Due to the characteristics of the Inception-v3, it outputs a vector of weights with 2,048 dimensions, which was used as a descriptor of the poster. As done in subsection \ref{subsec:audio}, transfer learning was applied in the weights initialization using the IMAGENET dataset \cite{imagenet}.

\subsection{Feature extraction from subtitle and synopsis (Phase 2)}
\label{subsec:text}

To represent textual content, which is given by subtitles and synopses in our dataset, we used two well-known text processing techniques: TF-IDF features extracted from N-grams \cite{damashek1995gauging}, and a representation learning approach based on Long Short-Term Memory (LSTM) \cite{hochreiter1997long}. Differently to the deep learning methods presented in subsections \ref{subsec:image}, \ref{subsec:audio} and \ref{subsec:poster}, LSTM was chosen to perform the classification of subtitles and synopsis. 

LSTMs are a special kind of Recurrent Neural Networks (RNN), explicitly designed to avoid the long-term dependency problem. We chose to use the LSTM because it has been used as one of the best RNN variations to solve the word-level language modeling context \cite{greff2017lstm}. Table \ref{lstm-params} shows the parameter settings used in the experiments with the LSTM algorithm in this study. We chose Word2vec \cite{word2vec} as the embedding representation because it has been presenting good results in several Natural Language Processing (NLP) tasks. Moreover, we defined the max features of this layer as 50,000. For every sample (i.e. synopsis or subtitle), we pass the first 300 words to the network, as shown in the parameters ``Num. of words'' and ``Input length''. We also defined the number of hidden nodes in the LSTM as 128, this decision defines the dimensionality of the output.

\begin{table}[htbp]
\centering
\caption{Parameter settings of the LSTM algorithm.}
\label{lstm-params}
\begin{tabular}{lc}
\hline
\textbf{Parameter} & \textbf{Value} \\ \hline
 Embedding & Word2Vec \\ 
 Preprocessing & Tokenizer \\ 
 Num. of Words (Tokenizer) & 300 \\ 
 Input Length & 300 \\ 
 Dropout Rate & 0.2 \\ 
 Optimizer & Adam \\ 
 Loss & Cross Entropy \\ 
 Max Features (Embedding) & 50,000 \\ 
 Activation & Softmax \\ 
 Dimensionality of the Output & 128 \\ 
 Epochs & 150 \\ \hline
\end{tabular}
\end{table}

To briefly describe the rationale that lies behind TF-IDF, we start from the concept of bag-of-word (BOW), which is one of the most used strategies to represent texts in the NLP field. The BOW representation counts the occurrences of words, without considering their position in the text, thus losing contextual information \cite{jurafsky2008speech}. Some contextual information can be obtained by counting word sequences, in a representation known as N-gram, the most well-known forms being: 1-gram, 2-gram, 3-gram, and 4-gram, which use sequences of one, two, three and four words, respectively. Although the use of N-grams is capable of aggregating more information about context when compared to BOW, it is still not enough to properly represent the context in some circumstances.

N-grams with high frequency in many documents may not be relevant to the learning process; however, n-grams with high frequency tend to be more relevant than less frequent ones \cite{jurafsky2008speech}. 
To emphasize important N-grams, considering both the frequencies in specific documents and all documents in the set of documents, it is common to divide each N-gram frequency (Term Frequency - TF) by its inverse frequency in other documents (Inverse Document Frequency - IDF). This procedure is usually called TF-IDF on BOW/N-grams. With the use of TF-IDF, the terms that appear in few documents are highlighted, but without disregarding their frequencies in a specific document.

Before generating the N-grams, we preprocessed the texts to remove special characters and perform stemming \cite{porter1980algorithm}, which reduces inflected words to their stem. These techniques were used mainly to reduce the N-grams dimensionality. Since we ranged N from one to four to create the N-grams used to calculate TF-IDF descriptors, we have a total of four feature sets for each data source (subtitles and synopsis).

\subsection{Summary of features and dimensionality reduction (Phases 2 and 3)}
\label{subsec:compress}

In this section, we describe some details about the features extracted from the different sources of information, and finally, we present the fusion rules used to integrate the unimodal results.

In total, 22 different types of feature sets were produced from different sources of information used in this work. 
The name assigned to the feature follows a pattern in which the data source and the feature are separated by the character ``-'' (e.g. ``DATA SOURCE-FEATURE'').
The features are organized in five different groups, according to the source of information used to produce them, as follows:

\begin{itemize}
    \item Features from the frames of the video trailers: 
    \begin{itemize}
        \item TRAILER-C3D: created using a three-dimensional CNN, as proposed in \cite{tran2015learning};
        \item TRAILER-LRCN: created using both CNN and LSTM for feature extraction;
        \item TRAILER-CTT-MMC: created using the CTT-MMC architecture, proposed by \cite{wehrmann2017movie};
        \item TRAILER-LBP: the LBP operator, proposed by \cite{ojala2001generalized} is used to extract texture descriptors of the frames. To combine the results obtained from different frames, the K-Means algorithm is applied;
         \item TRAILER-RGB: created using the histogram of the channels from the RGB color space. The K-Means algorithm is used to integrate the information taken from different frames.
    \end{itemize}
    \item Features from the audio content of the video trailers:
    \begin{itemize}
        \item AUDIO-MFCC: composed of coefficients obtained using MFCC, an audio content descriptor widely used to capture the timbral content of the sound \cite{hasan2004speaker};
        \item AUDIO-SSD: composed of SSD descriptors, which captures rhythmic and timbral information of the sound \cite{lidy2010suitability};
        \item AUDIO-SPEC-LBP: texture descriptors extracted from the time-frequency visual representation of the sound, as proposed in \cite{costa2012music};
        \item AUDIO-SPEC-INCv3: features learned by the Inception-v3 with transfer learning applied to the spectrogram image obtained from the sound.
    \end{itemize}
    \item Features from the movie posters:
    \begin{itemize}
        \item POSTER-LBP: texture features extracted from poster images using the LBP operator;
        \item POSTER-RGB: texture features extracted from poster images using the RGB histogram;
        \item POSTER-INCv3: features learned using the Inception-v3 with transfer learning applied to the image of the movie posters. 
    \end{itemize}
    \item Features from the movie subtitles:
    \begin{itemize}
        \item SUB-TFIDF-1: TF-IDF features extracted using a 1-gram model applied on the subtitles;
        \item SUB-TFIDF-2: TF-IDF features extracted using a 2-gram model applied on the subtitles;
        \item SUB-TFIDF-3: TF-IDF features extracted using a 3-gram model applied on the subtitles;
        \item SUB-TFIDF-4: TF-IDF features extracted using a 4-gram model applied on the subtitles;
        \item SUB-LSTM: created using the LSTM on the subtitles; 
    \end{itemize}
    \item Features from the movie synopses:
    \begin{itemize}
        \item SYN-TFIDF-1: TF-IDF features extracted using a 1-gram model applied on the synopses;
        \item SYN-TFIDF-2: TF-IDF features extracted using a 2-gram model applied on the synopses;
        \item SYN-TFIDF-3: TF-IDF features extracted using a 3-gram model applied on the synopses;
        \item SYN-TFIDF-4: TF-IDF features extracted using a 4-gram model applied on the synopses;
        \item SYN-LSTM: created using the LSTM on the synopses.
    \end{itemize}
\end{itemize}

An important aspect to consider when developing classifying systems concerns the dimensionality of the representations. The representations described in this section present particular issues that must be addressed before the classification phase. The sizes of the different feature vectors are far from uniform, which means that some descriptors are much bigger than others. Thus, in some cases, the configuration parameters of the classifiers to some feature vectors are not the most suitable for other feature vectors. By carefully analyzing the vectors produced from the different sources of information used here, we noticed that for some representations extracted from the text (such as 4-grams), the feature vectors can have an insurmountable size (about 51 million features). Therefore, they must be reduced to computational feasibility, and also to avoid problems related to the curse of dimensionality.

Proposed by \cite{compressiveSampling2}, compressive sampling is a method for dimensionality reduction that applies the concept of random projections for machine learning purposes. It was originally proposed to situations in which the signal has a sparse representation, so that it is possible to reconstruct the signal from a few random measurements. The mathematical details behind the method can be found in \cite{compressiveSampling3} and \cite{baraniuk2008simple}. After evaluating it empirically in our dataset, we observed that, for some feature vectors, the method is highly effective in reducing dimensionality while maintaining the hit rate. Therefore, we applied it to the text representations based on N-grams. Table \ref{tab:dimensionality} shows the dimensionality of feature vector obtained from the text used in this work, before and after the use of compressive sampling. 

\begin{table}[!htbp]
\caption{Dimensionality of features vectors obtained from text before and after applying compressive sampling.}\label{tab:dimensionality}
\centering
\begin{tabular}{c|lcc}
\hline
\textbf{Media source}&
\textbf{Feature type} &
\textbf{Before} &
\textbf{After}\\
\hline
\multirow{4}{*}{\textbf{Subtitle}}
& SUB-TFIDF-1 & 209,917    & 128 \\
& SUB-TFIDF-2 & 6,911,803  & 128 \\
& SUB-TFIDF-3 & 28,831,178 & 128 \\
& SUB-TFIDF-4 & 51,854,899 & 128 \\
\hline
\multirow{4}{*}{\textbf{Synopsis}}
& SYN-TFIDF-1 & 22,838  & 128 \\
& SYN-TFIDF-2 & 231,618 & 128 \\
& SYN-TFIDF-3 & 414,287 & 128 \\
& SYN-TFIDF-4 & 471,837 & 128 \\
\hline
\end{tabular}
\end{table}

\subsection{Resampling (Phase 4)}
\label{subsec:resampling}

Many researchers face problems related to unbalanced class distribution, which can make the task of learning an even greater challenge \cite{krawczyk2016learning}. 
Classifiers usually focus on minimizing the global error rate and, therefore, when dealing with unbalanced data, tend to benefit the most frequent classes, decreasing the prediction for infrequent classes and, consequently, affecting the overall performance by class.

Resampling techniques, which are the most common and widely used solution for the imbalance issue, can be subcategorized in oversampling and undersampling. While the first balances the dataset by creating new samples for the minority classes, the second aims at removing samples from the majority classes.

To deal with the imbalance in our movie genre classification schema, we included a resampling phase (represented by the Blue rectangle in Figure \ref{fig:overview}. We used two different resampling algorithms in this phase, namely the Multi-Label Synthetic Minority Over-sampling Technique (ML-SMOTE) \cite{charte2015mlsmote} and the Multi-Label Tomek Link (MLTL) \cite{pereira2019mltl}. 

ML-SMOTE was proposed by \cite{charte2015mlsmote} and we chose to use it because it is one of the most well-known resampling algorithms in the literature. Its main idea is to create synthetic samples combining the features of samples from the minority classes with interpolation techniques. In our experiments, the resize rate of ML-SMOTE was set to 25\% and a ranking label combination was chosen.

Proposed by \cite{pereira2019mltl}, MLTL is one of the most recently published resampling techniques in the literature. This undersampling algorithm detects and removes the so-called Tomek Links from the multi-label dataset. A pair of instances is a Tomek Link if they are the nearest neighbors but belong to different classes. Besides being a subsampling algorithm, the MLTL can also be applied in a post-process cleaning step for the ML-SMOTE algorithm. According to \cite{pereira2019mltl}, the reason to use it as a post-process cleaning step leans on the fact that, after applying ML-SMOTE, the class groups are usually not well defined, i.e., some instances from the majority class may be invading the minority class space or vice versa. Thus, MLTL may clean the feature space and smooth the edges between the classes.

\subsection {Classification algorithms and Multimodal integration (Phases 5 and 6)}
\label{subsec:integration}

In this subsection, we briefly describe the classification algorithms used in our experimental protocol and then describe how the classification was performed using representations created from the multimodal data previously described. 

According to \cite{tsoumakas2007multi}, we can group the existing methods for multi-label classification into two main categories: (1) problem transformation methods, and (2) algorithm adaptation methods. Problem transformation methods are those in which the multi-label classification problem is transformed into one or more single-label classification or regression problems, for which there is a huge bibliography of learning algorithms. Algorithm adaptation methods are those that extend specific learning algorithms to directly deal with multi-label data.

In this work, we chose to employ LSTM and Binary Relevance (BR), which are problem transformation methods. In the case of BR, we used Multilayer Perceptron (BR-MLP), Support Vector Machine (BR-SVM), and Decision Tree (BR-DT) as base classifiers. The BR method builds independent binary classifiers for each label ($l_i$). Each classifier maps the original dataset to a single binary label with values $l_i$, \~{}$l_i$. The classification of a new instance is given by the set of labels $l_i$ that are produced by the classifiers \cite{brinker2006unified}. In addition to the BR classifiers, we used the LSTM architecture to classify synopses and subtitles. 

We also experimented with the algorithm adaptation method ML-kNN. The Multi-Label k-Nearest Neighbors (ML-kNN) \cite{zhang2007ml} is an adaptation of the kNN lazy learning algorithm for multi-label data. In essence, ML-kNN uses the kNN algorithm independently for each label $l$: It finds the $k$ nearest examples to the test instance and considers those that are labeled at least with $l$ as positive and the rest as negative.

The multimodal integration itself was accomplished by late fusion. Late fusion strategies usually achieve good results in scenarios in which there is complementarity between the outputs of the classifiers involved in the fusion. In these cases, the classifiers do not make the same misclassification, so, when combined, they can help each other \cite{kittler1998combining}.

Late fusion strategies have this name because they combine the output of the classifiers, in opposition to early fusion strategies, which combine the feature vectors before executing the classification algorithm. In this sense, the combination is, in most cases, achieved through calculations in the predicted scores for each class involved in the problem.

Among the most used fusion strategies, we highlight the rules described as follows, originally introduced by \cite{kittler1998combining}, and adapted in this work for the multi-label scenario. 

\begin{itemize}
    \item Sum rule (Sum): 
    Corresponds to the sum of the scores provided by each classifier for each class. 
    \item Product rule (Prod): Corresponds to the product between the scores provided by each classifier for each class.
    \item Max rule (Max): Selects the highest probability score from each classifier.
\end{itemize}

The three aforementioned fusion rules were used to combine the scores provided by the classifiers.
In addition, we have introduced a threshold, so that classes with a score greater than or equal to 0.3 after the merger (Sum and Max rules) are considered qualified classes and are assigned to the pattern under classification. Due to the reduction in the probabilities values when the product rule is applied, its threshold was set as 0.01. Both thresholds were empirically adjusted. 

An important aspect regarding the multimodal integration concerns the criteria adopted to select the classifiers to be used in the fusion. We have tried two approaches: TOP-N, and BEST-ON-DATA. The former consists of selecting the N classifiers with the best overall performance based on the F-Score metric. The latter consists of using the best classifier for each kind of data source (i.e.: trailer frames, trailer audio, synopsis, poster, and subtitle), also considering the F-Score. In this case, we have always five classifiers to be combined. In total, we have generated 40 different classifiers using fusion (36 using TOP-N, and four using BEST-ON-DATA).

\section{Experimental protocol and results}
\label{sec:results}

Considering the number of sources of information, representations, and classification algorithms used in this work, it would not be appropriate to describe in this paper all the experimental results. This scenario could be even more complex if we consider the wide range of evaluation metrics for classification problems available in the literature. Therefore, we decided to organize and describe in this section the main results, emphasizing those situations where the best results were achieved. We also included an explanation regarding the chosen evaluation metrics. In addition, a complete description containing all our results was made available for those who are interested in it\footnote{https://drive.google.com/file/d/1uWbLhjAvHovFdqRIgeTJUkN5T2KovnoE/view?usp=sharing}. 
The experiments with the resampling strategy presented in subsection \ref{subsec:resampling} are also included in the results file, but they are not discussed in subsection \ref{subsec:results} because the use of this strategy did not improve the results.

\subsection{Evaluation metrics}
\label{sec:metrics}

There are several metrics proposed in the literature to evaluate multi-label classifiers. One of the main differences among them refers to the rigorousness in considering the classification result as a hit or a misclassification. In one extreme, we can point out ``Subset Accuracy'' as the most severe metric. Based on this metric, we have a hit only when the classifier correctly identifies the exact subset of labels assigned to the sample under classification. Other metrics, like Accuracy, Precision, and Recall \cite{herrera2016multilabel}, have a more relaxed criterion and thereby tends to provide higher hit rates. \cite{provost1998case} claim that the use of a single metric, accuracy for example, may lead to mistakes in the interpretation of results. Therefore, they recommend the use of AUC-ROC as a good metric for general purposes. On the other hand, \cite{davis2006relationship} claim that the AUC-ROC can lead to an optimistic view of the results when the dataset is unbalanced. Alternatively, the authors suggest the use of the AUC-PR metric, which is supposed to present similar results to AUC-ROC, but with a more realistic view reflected in the rates, better showing the space for improvement.

With this discussion in mind, we decided to use two metrics to evaluate the classifiers in our experiments: i) F-Score, because it is a harmonic mean between the values of recall and precision; and ii) AUC-PR, considering the argument of Davis and Goadrich aforementioned.

All experiments were conducted using 5-fold cross-validation. Therefore, for each combination of features (i.e. representation) and classifier, the training was performed five times using four folds, and the obtained model was tested on the remaining fold. The presented results correspond to the average of these five tests.

\subsection{Results}
\label{subsec:results}

Table \ref{tab:results} presents the best results, based on F-Score and AUC-PR, for each kind of representation of each of the different sources of information. 
In cases where the classifier that achieved the best F-Score rate does not match the one that provided the best AUC-PR rate, we presented both results.
As we can see, the best results were obtained using the representation SYN-LSTM (i.e. a deep learning method) both considering F-Score or AUC-PR metrics (shown in bold).

\begin{table}[!h]
\centering
\footnotesize
\caption{Best results for all the representations evaluated considering the F-Score and the AUC-PR.}\label{tab:results}
\begin{tabular}{llcc}
\hline
\multicolumn{4}{c}{\textbf{Trailer frames}}            \\ \hline
\textbf{Data source-Feature} & \textbf{Classifier} & \textbf{F-Score} & \textbf{AUC-PR} \\ \hline
TRAILER-C3D             & BR\_MLP      & 0.471          & 0.473           \\
TRAILER-LRCN            & BR\_MLP      & 0.292          & 0.395           \\
TRAILER-LRCN            & MLkNN         & 0.297          & 0.306                    \\
TRAILER-CTT-MMC         & BR\_MLP      & 0.317          & 0.362                    \\
TRAILER-LBP             & BR\_MLP      & 0.327          & 0.322                    \\
TRAILER-RGB             & BR\_MLP      & 0.318          & 0.285                    \\
TRAILER-RGB             & MLkNN         & 0.283          & 0.299                    \\ \hline \hline
\multicolumn{4}{c}{\textbf{Trailer audio}}               \\ \hline
\textbf{Data source-Feature} & \textbf{Classifier} & \textbf{F-Score} & \textbf{AUC-PR} \\ \hline
AUDIO-MFCC              & BR\_MLP      & 0.264          & 0.401             \\
AUDIO-MFCC              & MLkNN         & 0.312          & 0.315                    \\
AUDIO-SSD               & BR\_MLP      & 0.326          & 0.434                    \\
AUDIO-SPEC-LBP          & BR\_MLP      & 0.254          & 0.404                    \\
AUDIO-SPEC-LBP          & BR\_DT       & 0.312          & 0.189                    \\
AUDIO-SPEC-INCv3        & BR\_MLP      & 0.334          & 0.311                    \\ \hline \hline
\multicolumn{4}{c}{\textbf{Poster}}            \\ \hline
\textbf{Data source-Feature} & \textbf{Classifier} & \textbf{F-Score} & \textbf{AUC-PR} \\ \hline
POSTER-LBP              & BR\_MLP      & 0.223          & 0.385                    \\
POSTER-LBP              & BR\_DT       & 0.275          & 0.172                    \\
POSTER-RGB              & BR\_MLP      & 0.267          & 0.304                    \\
POSTER-RGB              & BR\_DT       & 0.276          & 0.172                    \\
POSTER-INCv3            & BR\_MLP      & 0.409          & 0.391                    \\ \hline \hline
\multicolumn{4}{c}{\textbf{Subtitle}}         \\ \hline
\textbf{Data source-Feature} & \textbf{Classifier} & \textbf{F-Score} & \textbf{AUC-PR}\\ \hline
SUB-TFIDF-1             & BR\_MLP      & 0.366          & 0.329                    \\
SUB-LSTM                & Deep Learning & 0.436          & 0.613                    \\ \hline \hline
\multicolumn{4}{c}{\textbf{Synopsis}}            \\ \hline
\textbf{Data source-Feature} & \textbf{Classifier} & \textbf{F-Score} & \textbf{AUC-PR}  \\ \hline
SYN-TFIDF-1             & BR\_MLP      & 0.288          & 0.266                    \\
SYN-TFIDF-1             & MLkNN         & 0.225          & 0.285                    \\
SYN-LSTM                & Deep Learning & \textbf{0.488} & \textbf{0.678}           \\ \hline
\end{tabular}
\end{table}

Table \ref{tab:results-top-n} shows the best results considering the fusion of the TOP-N classifiers, with N ranging from 2 to 10. The TOP-N classifiers were selected from the 10 best classifiers (in terms of F-Score rate) among all possible classifiers (i.e., all possible combinations of representation and classifier algorithm). These 10 best classifiers are described in Table \ref{tab:best-10-fscore}. Similarly to Table \ref{tab:results}, in Table \ref{tab:results-top-n} we have duplicated the rows of TOP-N where the best results with AUC-PR and F-Score were obtained using different fusion rules. 
Again, the best results are in bold.

\begin{table}[!h]
\centering
\small
\caption{Best TOP-N fusion results with N ranging from 2 to 10 considering the F-Score and the AUC-PR. ``Proba'' refers to the probabilities, and ``Pred'' refers to the predictions produced by the classifiers.}\label{tab:results-top-n}
\begin{tabular}{clcc}
\hline
\textbf{Strategy} & \textbf{Fusion Method} & \textbf{F-Score} & \textbf{AUC-PR}  \\ \hline
TOP-2             & Proba\_Prod           & \textbf{0.628}   & 0.664                          \\
TOP-3             & Proba\_Prod           & 0.326            & 0.636                          \\
TOP-3             & Pred\_Sum             & 0.577            & 0.483                 \\
TOP-4             & Proba\_Sum            & 0.527            & 0.566                          \\
TOP-4             & Proba\_Prod           & 0.229            & 0.636                          \\
TOP-5             & Proba\_Prod           & 0.099            & \textbf{0.673}                 \\
TOP-5             & Pred\_Sum             & 0.563            & 0.539                          \\
TOP-6             & Proba\_Sum            & 0.544            & 0.619                          \\
TOP-6             & Pred\_Sum             & 0.590            & 0.555                          \\
TOP-7             & Proba\_Sum            & 0.549            & 0.602                          \\
TOP-8             & Proba\_Sum            & 0.555            & 0.607                          \\
TOP-8             & Pred\_Sum             & 0.569            & 0.564                          \\
TOP-9             & Proba\_Sum            & 0.532            & 0.606                          \\
TOP-9             & Pred\_Sum             & 0.578            & 0.567                          \\
TOP-10            & Proba\_Sum            & 0.531            & 0.603                          \\
TOP-10            & Proba\_Sum            & 0.540            & 0.568                          \\ \hline
\end{tabular}
\end{table}

\begin{table}[]
\centering
\caption{Ten best unimodal classifiers considering the F-Score.}\label{tab:best-10-fscore}
\begin{tabular}{cllc}
\hline
\# & Data source-Feature          & Classifier    & F-Score \\ \hline
1  & SYN-LSTM         & Deep Learning & 0.488    \\
2  & TRAILER-C3D      & BR\_MLP      & 0.471    \\
3  & TRAILER-C3D      & MLkNN         & 0.457    \\
4  & TRAILER-C3D      & BR\_SVM      & 0.457    \\
5  & SUB-LSTM         & Deep Learning & 0.436    \\
6  & POSTER-INCv3     & BR\_MLP      & 0.409    \\
7  & TRAILER-C3D      & BR\_DT       & 0.402    \\
8  & SUB-TFIDF-1      & BR\_MLP      & 0.366    \\
9  & POSTER-INCv3     & MLkNN         & 0.355    \\
10 & AUDIO-SPEC-INCv3 & BR\_MLP      & 0.334    \\ \hline
\end{tabular}
\end{table}

The best results obtained by the unimodal classifiers created from each data source in isolation are described in Table \ref{tab:best-on-data-classifiers}. Conversely, Table \ref{tab:results-best-on-data} shows the results obtained by fusing the classifiers with best F-Scores from each different source of information. We called these latter ``best-on-data''.

\begin{table}[]
\centering
\caption{Best classifiers for each kind of data source considering F-Score.}\label{tab:best-on-data-classifiers}
\begin{tabular}{lllc}
\hline
\textbf{Data source} & \textbf{Feature} & \textbf{Classifier} & \textbf{F-Score} \\ \hline
Trailer frames     & TRAILER-C3D      & BR\_MLP            & 0.471             \\
Trailer audio      & AUDIO-SPEC-INCv3 & BR\_MLP            & 0.334             \\
Poster             & POSTER-INCv3     & BR\_MLP            & 0.409             \\
Subtitle           & SUB-LSTM         & Deep Learning       & 0.436             \\
Synopsis           & SYN-LSTM         & Deep Learning       & 0.488             \\ \hline
\end{tabular}
\end{table}

\begin{table}[]
\centering
\caption{BEST-ON-DATA results.}\label{tab:results-best-on-data}
\begin{tabular}{ccc}
\hline
\textbf{Fusion Method} & \textbf{F-Score} & \textbf{AUC-PR} \\ \hline
Proba\_Sum            & 0.558             & \textbf{0.622}                   \\
Proba\_Max            & 0.495             & 0.465                   \\
Proba\_Prod           & 0.263             & 0.547                            \\
Pred\_Sum             & \textbf{0.574}    & 0.534                            \\ \hline
\end{tabular}
\end{table}

\begin{figure}[!h]
    \centering
    \includegraphics[width=1.0\textwidth]{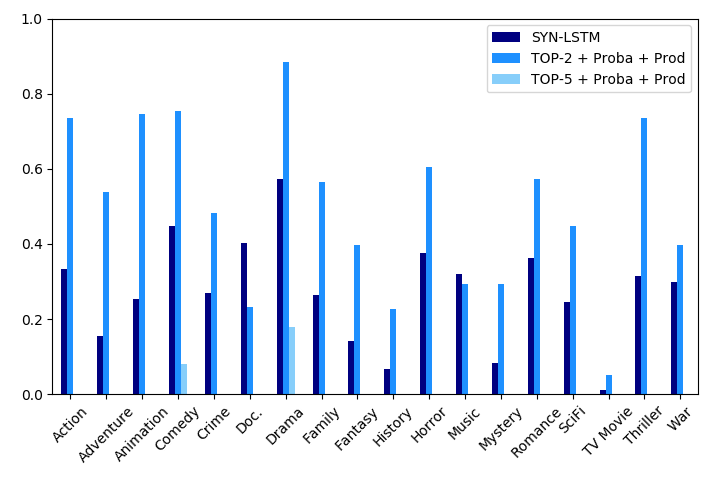}
    \caption{Hit per genre considering the best classifiers with and without fusion.}
    \label{fig:hit-per-genre}
\end{figure}

Figure \ref{fig:hit-per-genre} shows the recall per label considering the best classifiers with and without fusion, based on both F-Score and AUC-PR metrics. As can be observed, the values for ``TOP-5\_Proba\_Prod'' are zero for all the genres, except for Comedy and Drama. We conjecture that this is probably due to the low misclassification rate, accompanied by a low hit rate caused by the Prod rule. However, this combination maintains its consistency in the precision-recall curve.

In the next section, we discuss the results presented here. We also present some statistical tests comparing the best results obtained with or without fusion.

\section{Discussion}
\label{sec:discussion}

To analyze the results from different perspectives, we decided to guide our discussion in the search for answers to the following questions: (1) Which data source/representation provided the best and worst individual results? (2) Has the fusion contributed to improve the results? (3) Which classifier performed better as a whole? (4) Which classifier algorithm, and with which representations, contributed to the best fusions in terms of performance? (5) Which movie genres are easier/harder to identify?

One of the first questions that come to mind about our results is related to the data source/representation which provides the best and worst individual results (question 1). Looking at Table \ref{tab:results}, we observe that the best rate, in terms of both F-Score and AUC-PR, was achieved by a deep learning method, i.e. LSTM. The best result was obtained on the synopses, and the second-best was obtained on the subtitles considering the AUC-PR metric. This evinces the strength of the features created by deep learning methods on textual data in the application domain investigated here.
It is important to mention that, considering the F-Score metric, the second-best rate was also obtained using this representation. Considering all the results obtained from each source of information described in Table \ref{tab:results}, the worst performance in terms of F-Score was obtained using trailer audio, with a top result of 0.334. Considering the AUC-PR, the worst result regarding the different types of data sources remains on the use of poster, with a top result of 0.391 (POSTER-INCv3).


Regarding the fusion strategies experimented in this work (question 2), we can highlight the fusion performed between TOP-N classifiers. The best F-Score rate, considering all results presented in this paper, was obtained by fusing the two best classifiers considering their individual results. In terms of AUC-PR, the best result was obtained using the five best classifiers individually. Among the five best individual classifiers, we can find classifiers created using data from synopses, trailer frames, and subtitles. This scenario confirms one of the main hypotheses investigated in this work, concerning the complementarity between different classifiers/data sources.

We have also performed a statistical test in order to properly compare the best results obtained with and without fusion for each metric, aiming to check if there is a significant statistical difference between those results, and indicate that the fusion strategies improve the results (question 2). For the results without fusion, we used the SYN-LSTM classifier, both for F-Score and AUC-PR. In turn, the strategy TOP-2 with the ``Prod\_Proba'' fusion rule was used to evaluate the results using fusion considering the F-Score metric. For the AUC-PR metric, the strategy TOP-5 with the rule ``Prod\_Proba'' was used. As Hypothesis One, we used the \cite{wilcoxon1992individual} two-side test to verify whether or not the hypothesis of the results with and without fusion are equal. We also presented a Hypothesis Two using the Wilcoxon less hypothesis test to check if the mean of the difference between the results with and without fusion is negative (the results without fusion cannot be higher than the results with fusion). As described in Table \ref{tab:statistics}, we can discard the Hypothesis One with a statistical confidence level of 5\%,  and consider true the Hypothesis Two with a statistical confidence level of 5\% considering the F-Score metric. The AUC-PR did not present statistical relevance considering a confidence value of 5\%. These tests indicate that the fusion of classifiers obtained using different kinds of data source improve the result when analyzing the F-Score metric (Question 2).

\begin{table}[]
\centering
\caption{Statistical tests comparing the best results obtained with and without fusion.}\label{tab:statistics}
\begin{tabular}{ccc}
\hline
Metric  & Wilcoxon & Wilcoxon ``Less'' Hypothesis \\ \hline
F-Score & 0.043    & 0.022 \\
AUC-PR  & 0.893    & 0.554 \\ \hline
\end{tabular}
\end{table}

Regarding the classifier algorithms evaluated here, by looking at Table \ref{tab:results}, we can observe that the BR\_MLP classifier is the one with the biggest number of occurrences, fourteen in total. This algorithm also appears at least once for each different kind of data source. 
Those affirmations indicate that the BR\_MLP classifier performed better in comparison with the other classifiers when looking in a wide scenario (question 3).
Despite this impressive number of occurrences, we need to point out that the LSTM deep learning strategy has figured as the best one in terms of hit rate, both considering F-Score and AUC-PR, especially when applied on the text of the synopsis. Also appearing as one of the most important classifiers in the fusion stage (question 4).

Figure \ref{fig:hit-per-genre} lets us evaluate the results individually obtained by the 18 labels of the database (question 5). By analyzing those results, we can observe that five classes present hit rates above 70\%: Drama, Comedy, Animation, Action, and Thriller. In all these cases, the results were obtained with TOP-2+Pod+Proba. Considering this same classifier, the worst results appear for the genre TV Movie. However, if we consider the results obtained with TOP-5, there are several other labels with even worse results (available on the weblink with complete results).
This presents an important aspect of the dataset, when the genres with most samples (i.e. Drama, Comedy, Animation, Action, and Thriller) are easier to classify, and the genre with fewer samples (i.e. TV movie) is the hardest one to predict. It is worth remembering that we tried a resampling strategy aiming to prevent the impacts of the dataset imbalance, but even in these cases we did not get better results.


In summary, the data source that best performs individually was the synopsis with the LSTM classifier (Question 1). On the other hand, the poster and the trailer audio presented the worst results (Question 1). To compare the improvements achieved when using late fusion strategies, we applied the Wilcoxon statistical test. For the F-Score metric, we checked with a confidence interval of 5\% that the result obtained with the fusion strategy is higher than the conventional approach. The test indicates that the use of late fusion improves the classification performance (Question 2). The classifier that mostly occurs on the best results for the features is the BR\_MLP, showing to be a consistent and reliable classifier (Question 3). The LSTM classifier figured as the one that most contributed to the fusion, presenting the two best results for the AUC-PR metric (Question 4). Regarding the classification of each genre individually, genres with more samples are easiest to classify (i.e. Drama and Action) (Question 5). Therefore, the hardest genre to classify is also the one with fewer samples (i.e. TV movie) (Question 5).

\section {Concluding remarks and future works}
\label{sec:conclusion}

In this work, we addressed the multimodal movie genre classification as a multi-label classification problem. As far as we know, this is the most comprehensive study done in this scenario. To perform the classification, we investigated the use of five different sources of information, namely trailer audio, trailer frames, synopses, subtitles, and posters. We carefully designed and curated a dataset to support the development of the experimental protocol, and made it available to the research community, in order to allow that other researchers can fairly compare their proposals on the same task. 

We created classifiers using different representations based on the different sources of data, and they were evaluated both individually, and combined with each other by late fusion. The results were reported using F-Score and AUC-PR metrics, and in both cases, the best individual result was obtained using an LSTM representation created from the synopses. Regarding the classifiers combined by fusion, the best F-Score rate, of 0.628, was achieved when the SYN-LSTM (i.e. LSTM created using synopses) was combined with TRAILER-C3D (i.e. CNN created using TRAILER frames). Considering the AUC-PR metric, the best rate, of 0.673, was obtained using SYN-LSTM, TRAILER-C3D, and also SUB-LSTM (i.e. LSTM created from subtitles). These results corroborate the existence of complementarity between classifiers created using different sources of information in the classification task addressed here. These results also confirm the success of deep learning strategies to perform multimedia classification, already investigated by the authors on music classification.

In the future, we intend to investigate the use of optimization methods (e.g. Particle Swarm Optimization (PSO), Genetic Algorithm (GA), among others) to conduct the search for the best classifiers combination, aiming at finding those who provide the best results. We also plan to evaluate the use of dynamic classifiers selection techniques and the use of segment selection methods aiming at automatically identifying the most promising parts of the movie trailer to perform the genre classification.

\section*{Acknowledgements}

We thank the Brazilian Research Support Agencies: CNPq - National Council for Scientific and Technological Development (grants \#156956/2018-7 and \#164837/2018-3), CAPES - Coordination for the Improvement of Higher Education Personnel and FA - Araucária Foundation for their financial support for their financial support. We also gratefully acknowledge the support of NVIDIA Corporation with the donation of a Titan XP GPU used in this research.

\bibliographystyle{unsrt}
\bibliography{mybibfile}

\end{document}